\documentclass{llncs}

\usepackage{llncsdoc}

\usepackage{amsmath}
\usepackage{amsfonts}
\usepackage[dvipsnames]{xcolor}
\usepackage{graphicx}

\title{Towards Grounding Conceptual Spaces in Neural Representations\thanks{The final version of this paper is published at \url{http://ceur-ws.org/Vol-2003/}.}}
\author{Lucas Bechberger\thanks{Corresponding author, ORCID: 0000-0002-1962-1777} \and Kai-Uwe K\"uhnberger}
\institute{Institute of Cognitive Science, Osnabr\"uck University, Osnabr\"uck, Germany \email{lucas.bechberger@uni-osnabrueck.de}, \email{kai-uwe.kuehnberger@uni-osnabrueck.de}}

\begin{document}
\maketitle


\begin{abstract}
The highly influential framework of conceptual spaces provides a geometric way of representing knowledge. It aims at bridging the gap between symbolic and subsymbolic processing. Instances are represented by points in a high-dimensional space and concepts are represented by convex regions in this space. In this paper, we present our approach towards grounding the dimensions of a conceptual space in latent spaces learned by an InfoGAN from unlabeled data.
\end{abstract}

\section{Introduction}
\label{Intro}

The cognitive framework of conceptual spaces \cite{Gardenfors2000,Gardenfors2014} attempts to bridge the gap between symbolic and subsymbolic AI by proposing an intermediate conceptual layer based on geometric representations.
A conceptual space is a high-dimensional space spanned by a number of quality dimensions representing interpretable features. Convex regions in this space correspond to concepts. Abstract symbols can be grounded by linking them to concepts in a conceptual space whose dimensions are based on subsymbolic representations.

The framework of conceptual spaces has been highly influential in the last 15 years within cognitive science and cognitive linguistics \cite{Douven2011,Fiorini2013,Warglien2012}. It has also sparked considerable research in various subfields of artificial intelligence, ranging from robotics and computer vision \cite{Chella2005,Chella2001,Chella2003} over the semantic web and ontology integration \cite{Adams2009a,Dietze2008} to plausible reasoning \cite{Derrac2015,Schockaert2011}.\\

Although this framework provides means for representing concepts, it does not consider the question of how these concepts can be learned from mostly unlabeled data. Moreover, the framework assumes that the dimensions spanning the conceptual space are already given a priori. In practical applications of the framework, they thus often need to be handcrafted by a human expert.

In this paper, we argue that by using neural networks, one can automatically extract the dimensions of a conceptual space from unlabeled data. We propose that latent spaces learned by an InfoGAN \cite{Chen2016} (a special class of Generative Adversarial Networks \cite{Goodfellow2014}) can serve as domains in the conceptual spaces framework. We further propose to use a clustering algorithm in these latent spaces in order to discover meaningful concepts.

The remainder of this paper is structured as follows:
Section \ref{CS} presents the framework of conceptual spaces and Section \ref{RepresentationLearning} introduces the InfoGAN framework. In Section \ref{DomainGrounding}, we present our idea of combining these two frameworks. Section \ref{Example} gives an illustrative example and Section \ref{Conclusion} concludes the paper.

\section{Conceptual Spaces}
\label{CS}

A conceptual space \cite{Gardenfors2000} is a high-dimensional space spanned by so-called ``quality dimensions''. Each of these dimensions represents an interpretable way in which two stimuli can be judged to be similar or different. Examples for quality dimensions include temperature, weight, time, pitch, and hue.
A domain is a set of dimensions that inherently belong together. Different perceptual modalities (like color, shape, or taste) are represented by different domains. The color domain for instance can be represented by the three dimensions hue, saturation, and brightness.\footnote{Of course, one can also use other color spaces, e.g., the CIE L*a*b* space.} Distance within a domain is measured by the Euclidean metric.

The overall conceptual space is defined as the product space of all dimensions. Distance within the overall conceptual space is measured by the Manhattan metric of the intra-domain distances.
The similarity of two points in a conceptual space is inversely related to their distance -- the closer two instances are in the conceptual space, the more similar they are considered to be.

The framework distinguishes properties like ``red'', ``round'', and ``sweet'' from full-fleshed concepts like ``apple'' or ``dog'': Properties are represented as regions within individual domains (e.g., color, shape, taste), whereas full-fleshed concepts span multiple domains. Reasoning within a conceptual space can be done based on geometric relationships (e.g., betweenness and similarity) and geometric operations (e.g., intersection or projection). 

Recently, Balkenius \& G\"ardenfors \cite{Balkenius2016} have argued that population coding in the human brain can give rise to conceptual spaces. They discuss the connection between neural and conceptual representations from a neuroscience/psychology perspective, whereas we take a machine learning approach in this paper.

\section{Representation Learning with InfoGAN}
\label{RepresentationLearning}

Within the research area of neural networks, there has been some substantial work on learning compressed representations of a given feature space. Bengio et al. \cite{Bengio2013} provide a thorough overview of different approaches in the representation learning area. They define representation learning as ``learning representations of the data that make it easier to extract useful information when building classifiers or other predictors''. We will focus our discussion here on one specific approach that is particularly fitting to our proposal, namely InfoGAN \cite{Chen2016}. InfoGAN is an extension of the GAN (Generative Adversarial Networks) framework \cite{Goodfellow2014} which has been applied to a variety of problems (e.g., \cite{Durugkar2016,Radford2015,Wu2016,Zhao2016,Zhu2017}). We first describe the original GAN framework before moving on to InfoGAN.\\

\begin{figure}[t]
\centering
\includegraphics[width = 1.0\columnwidth]{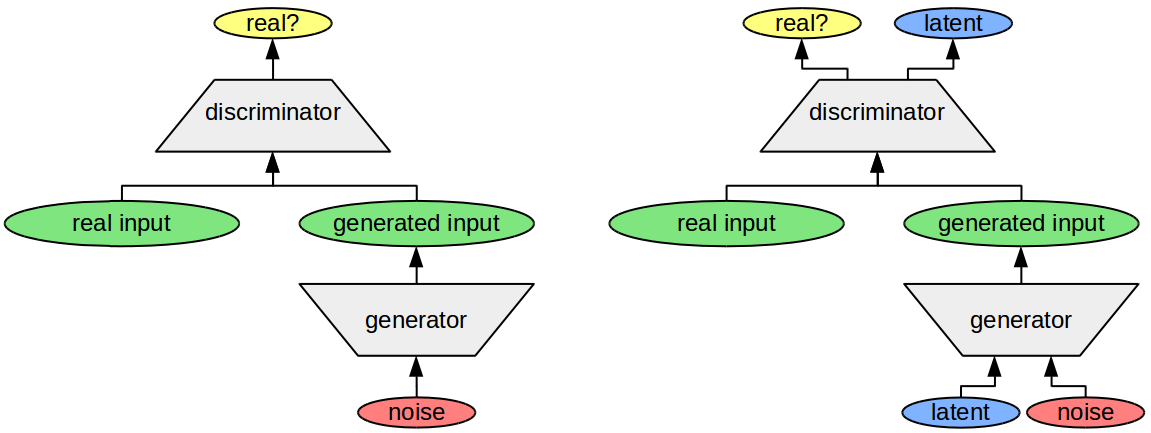} 
\caption{Left: Illustration of a GAN. Right: Illustration of an InfoGAN.}
\label{fig:RepresentationLearning}
\end{figure}

The GAN framework (depicted in the left part of Figure \ref{fig:RepresentationLearning}) consists of two networks, the generator and the discriminator. The generator is fed with a low-dimensional vector of noise values. Its task is to create high-dimensional data vectors that have a similar distribution as real data vectors taken from an unlabeled training set. The discriminator receives a data vector that was either created by the generator or taken from the training set. Its task is to distinguish real inputs from generated inputs. Although the discriminator is trained on a classification task, the overall system works in an unsupervised way. The overall architecture can be interpreted as a two-player game: The generator tries to fool the discriminator by creating realistic inputs and the discriminator tries to avoid being fooled by the generator. When the GAN framework converges, the discriminator is expected to make predictions only at chance level and the generator is expected to create realistic data vectors. Although the overall framework works quite well, the dimensions of the input noise vector are usually not interpretable.

Chen et al. \cite{Chen2016} have extended the original framework by introducing latent variables: In the InfoGAN framework (shown in the right part of Figure \ref{fig:RepresentationLearning}), the generator receives an additional input vector. The entries of this vector are values of latent random variables, selected based on some probability distribution that was defined a priori (e.g., uniform or Gaussian). The discriminator has the additional task to reconstruct these latent variables.\footnote{This introduces a structure similar to an autoencoder (with the latent variables as input/output and the generated data vector as hidden representation).} Chen et al. argue that this ensures that the mutual information between the latent variable vector and the generated data vector is high. They showed that after training an InfoGAN, the latent variables tend to have an interpretable meaning. For instance, in an experiment on the MNIST data set, the latent variables corresponded to type of digit, digit rotation and stroke thickness. InfoGANs can thus provide a bidirectional mapping between observable data vectors and interpretable latent dimensions: One can both extract interpretable dimensions from a given data vector and create a data vector from an interpretable latent representation.

\section{Using Representation Learning to Ground Domains}
\label{DomainGrounding}

For some domains of a conceptual space, a dimensional representation is already available. For instance, the color domain can be represented by the three-dimensional HSB space. 
For other domains, it is however quite unclear how to represent them based on a handful of dimensions. One prominent example is the shape domain: To the best of our knowledge, there are no widely accepted dimensional models for describing shapes.

We propose to use the InfoGAN framework in order to learn such a dimensional representation based on an unlabeled data set: Each of the latent variables can be interpreted as one dimension of the given domain of interest. For instance, the latent variables learned on a data set of shapes can be interpreted as dimensions of the shape domain.
Three important properties of domains in a conceptual space are the following: interpretable dimensions, a distance-based notion of similarity, and a geometric way of describing semantic betweenness. We think that the latent space of an InfoGAN is a good candidate for representing a domain of a conceptual space, because it fulfills all of the above requirements:

As described before, Chen et al. \cite{Chen2016} found that the individual latent variables tend to have an interpretable meaning. Although this is only an empirical observation, we expect it generalize to other data sets and thus to other domains.

Moreover, the \textit{smoothness assumption} used in representation learning (cf. \cite{Bengio2013} and \cite[Ch. 15]{Goodfellow2016}) states that points with small distance in the input space should also have a small distance in the latent space. This means that a distance-based notion of similarity in the latent space is meaningful.

Finally, Radford et al. \cite{Radford2015} found that linear interpolations between points in the latent space of a GAN correspond to a meaningful ``morph'' between generated images in the input space. This indicates that geometric betweenness in the latent space can represent semantic betweenness.\\

There are two important hyperparameters to the approach of grounding domains in InfoGANs: The number of latent variables (i.e., the dimensionality of the learned domain) and the type of distribution used for the latent variables (e.g., uniform vs. Gaussian). Note that one would probably aim for the lowest-dimensional representation that still describes the domain sufficiently well.\\

Finally, we would like to address a critical aspect of this proposal: How can one make sure that the representation learned by the neural network only represents information from the target domain (e.g., shape) and not anything related to other domains (e.g., color)?
In our opinion, there are two complementary methods to ``steer'' the network towards the desired representation:

The first option consists of selecting only such inputs for the training set that do not exhibit major differences with respect to other domains. For instance, a training set for the shape domain should only include images of shapes that have the same color (e.g., black shape on white ground). If there is only very small variance in the data set with respect to other domains (e.g., color), the network is quite unlikely to incorporate this information into its latent representation.

The second option concerns modifications of the network's loss function: One could for instance introduce an additional term into the loss function which measures the correlation between the learned latent representation and dimensions from other (already defined) domains. This would cause a stronger error signal if the network starts to re-discover already known dimensions from other domains and therefore drive the network away from learning redundant representations.\\

A simple proof of concept implementation for the shape domain could be based on a data set of simple 2D shapes (circles, triangles, rectangles, etc.) in various orientations and locations. For a more thorough experiment, one could for instance use ShapeNet\footnote{\url{https://www.shapenet.org/}} \cite{Chang2015}, a data base of over 50,000 3D models for more than 50 categories of objects. One could render these 3D models from various perspectives in order to get 2D inputs (for learning to represent 2D shapes) or work on a voxelized 3D input (for learning representations of 3D shapes). 

\section{An Illustrative Example}
\label{Example}

\begin{figure}[t]
\centering
\includegraphics[width = 0.75\columnwidth]{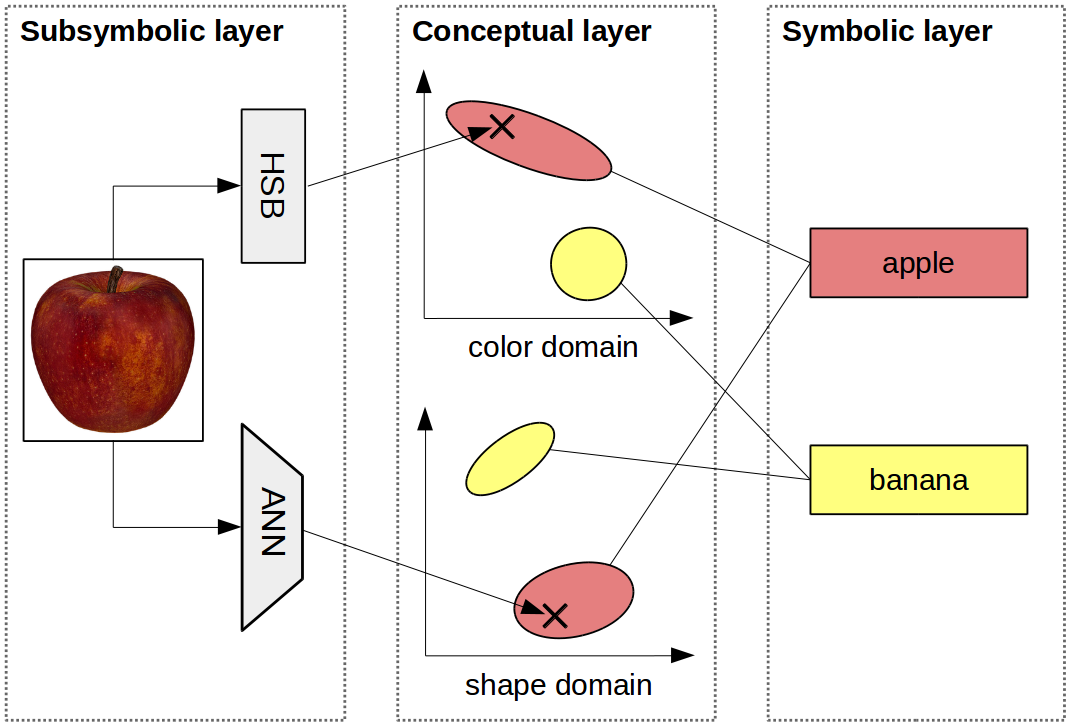} 
\caption{Illustration of the envisioned overall system. Perceptions from the subsymbolic layer are transformed to points in different domains by feature extractors that are either handcrafted (e.g., HSB for colors) or trained neural networks (e.g., for the shape domain). Each concept is described by one region per domain.}
\label{fig:Architecture}
\end{figure}

Figure \ref{fig:Architecture} illustrates a simplified example of our envisioned overall system. Here, we consider only two domains: color and shape. Color can be represented by the HSB space using the three dimensions hue, saturation and brightness. This is an example for a hard-coded domain. The representation of the shape domain, however, needs to be learned. The artificial neural network depicted in Figure \ref{fig:Architecture} corresponds to the discriminator of an InfoGAN trained on a data set of shapes.\\

Let us consider two example concepts: The concept of an apple can be described by the ``red'' region in the color domain and the ``round'' region in the shape domain. The concept of a banana can be represented by the ``yellow'' region in the color domain and the ``cylindric'' region in the shape domain.

If the system makes a new observation (e.g., an apple as depticted in Figure \ref{fig:Architecture}), it will convert this observation into a point in the conceptual space. For the color domain, this is done by a hard-coded conversion to the HSB color space. For the shape domain, the observation is fed into the discriminator and its latent representation is extracted, resulting in the coordinates for the shape domain. Now in order to classify this observation, the system needs to check whether the resulting data point is contained in any of the defined regions. If the data point is an element of the apple region in both domains (which is the case in our example), this observation should be classified as an apple. If the data point is an element of the banana region, the object should be classified as a banana. 

Based on a new observation, the existing concepts can also be updated: If the observation was classified as an apple, but it is not close to the center of the apple region in one of the domains, this region might be enlarged or moved a bit, such that the observed instance is better matched by the concept description. If the observation does not match any of the given concepts at all, even a new concept might be created. This means that concepts cannot only be applied for classification, but they can also be learned and updated. Note that this can take place without explicit label information, i.e., in an unsupervised way. Our overall reserach goal is to develop a clustering algorithm that can take care of incrementally updating the regions in such a conceptual space. 

Please note that the updates considered above only concern the connections between the conceptual and the symbolic layer. The connections between the subsymbolic and the conceptual layer remain fixed. The neural network thus only serves as a preprocessing step in our approach: It is trained before the overall system is used and remains unchanged afterwards. Simultaneous updates of both the neural network and the concept description might be desirable, but would probably introduce a great amount of additional complexity.

\section{Conclusion and Future Work}
\label{Conclusion}

In this paper, we outlined how neural representations can be used to ground the domains of a conceptual space in perception. This is especially useful for domains like shape, where handcrafting a dimensional representation is difficult. We argued that the latent representations learned by an InfoGAN have suitable properties for being combined with the conceptual spaces framework. In future work, we will implement the proposed idea by giving a neural grounding to the domain of simple 2D shapes. Furthermore, we will devise a clustering algorithm for discovering and updating conceptual representations in a conceptual space.

%

\end{document}